\pdfoutput=1

\documentclass[11pt]{article}

\usepackage[final]{acl}

\usepackage{times}
\usepackage{latexsym}
\usepackage{booktabs}
\usepackage{amsthm}
\newtheorem{theorem}{Theorem}
\usepackage[T1]{fontenc}

\usepackage[utf8]{inputenc}
\usepackage{multirow} 
\usepackage{enumitem}
\usepackage{tablefootnote}
\usepackage{marvosym}
\usepackage{amsfonts}
\usepackage{amsmath}
\usepackage{microtype}
\usepackage{adjustbox}
\usepackage{inconsolata}

\usepackage{graphicx}

\title{Each graph is a new language: Graph Learning with LLMs}

\author{Huachi Zhou$\dagger$, Jiahe Du$\dagger$, \textbf{Chuang Zhou}, \\
\textbf{Chang Yang}, \textbf{Yilin Xiao}, \textbf{Yuxuan Xie}, \textbf{Xiao Huang}\Thanks{Xiao Huang is the corresponding author.}\\
The Hong Kong Polytechnic University, China\\ 
\{huachi.zhou, jiahe.du, chuang-qqzj.zhou, chang.yang\}@connect.polyu.hk; \\ \{yilin.xiao, yuxuan.xie\}@connect.polyu.hk; xiaohuang@comp.polyu.edu.hk \\}
\begin{document}
\maketitle
\begingroup\def\thefootnote{$\dagger$}\footnotetext{Huanchi Zhou and Jiahe Du are co-first authors. Both authors contributed equally to this research.}\endgroup
\begin{abstract}
Natural language has been extensively used for modeling text-attributed graphs with LLMs. Natural language is used to describe the graph for LLMs to understand or serve as component of the graph, e.g., textual attributes for embedding generation. However, natural language is inherently redundant and unstructured, making it unsuitable for modeling high-order neighbors with LLMs. Specifically, (i) graph descriptions become verbose, overwhelming LLMs, and (ii) only relying on attribute embeddings limits LLM's ability to capture the adequate graph structural information. These limitations make it difficult to model graphs both concisely and adequately using sole natural language with LLMs.

Inspired by the observation that LLMs pre-trained on one language can achieve exceptional performance on another with minimal additional training, we propose \textbf{G}raph-\textbf{D}efined \textbf{L}anguage for \textbf{L}arge \textbf{L}anguage \textbf{M}odel (GDL4LLM). This novel framework enables LLMs to transfer their powerful language understanding capabilities to graph-structured data. GDL4LLM translates the graph into a graph language corpus instead of graph descriptions and pre-trains LLMs on this corpus to adequately understand the graph. This corpus represents the subgraph centered around target nodes concisely with only a few tokens during fine-tuning on downstream tasks. By treating the graph as a new language, GDL4LLM enables LLMs to model text-attributed graph adequately and concisely. Extensive experiments on three real-world datasets demonstrate that GDL4LLM outperforms description-based and embedding-based baselines by efficiently modeling different orders of neighbors.
\end{abstract}

\section{Introduction}

Text-attributed graphs have become essential data representations in various domains, such as social networks and citation networks~\cite{huang2019graph,guo2024graphedit}. One of the key components of these graphs is natural language text, which serves as an attribute and a label associated with nodes. For example, in a citation network, nodes represent papers, and their text attributes could include abstracts, while labels correspond to paper categories. To accurately predict a node's label, pioneering studies have shown that a node's class is dependent on its attributes and those of neighboring nodes in the graph~\cite{cook2006mining}.

To capture this dependency on graph, recent approaches have aimed to leverage LLMs in modeling text-attributed graphs. LLMs, such as GPTs~\cite{achiam2023gpt}, have demonstrated exceptional text manipulation capabilities.~\cite{hong2024next,zhang2025survey}. To adapt these capabilities to graph-structured data, natural language plays a key role, and associated approaches could be categorized into two main classes: (i) describing the graph in natural language to enable LLMs to comprehend, by enumerating nodes and their connections ~\cite{guan2024langtopo,chen2024llaga}; and (ii) using LLMs to embed components of the graph, i.e., attributes, and aggregating attribute embeddings to encode graph-level information~\cite{huang2022going,zhao2022learning}.

However, natural
language is inherently redundant and unstructured, making it unsuitable for modeling high-order neighbors of target nodes with LLMs: (i) \textbf{Graph descriptions become excessively verbose and overwhelm LLMs.} Graph structures are inherently complex, and intricate connections among nodes often become convoluted when translated into natural language descriptions~\cite{fatemitalk}. These convoluted descriptions make it difficult for LLMs to reason effectively and prioritize the high-order neighbors relevant to the target nodes. (ii) \textbf{Only relying on attribute embeddings limits the ability of LLM to capture
adequate structural information.} Graph structure is not deeply involved in the LLM embedding generation. To capture structure information, GNN aggregation mixes embeddings with those of neighbors but risks oversmoothing, especially when attempting to reach high-order neighbors~\cite{chen2024exploring}. And due to excessive computational overhead, LLMs' optimization is decoupled from the aggregation. This decoupling limits LLMs' ability to effectively model high-order neighbors and assist in integrating rich structural information into embeddings~\cite {huang2024can}.

It is challenging to model high-order neighbors adequately and concisely with LLMs: (i) \textbf{LLM Constraints in Modeling Graphs}: Unlike GNNs, which leverage built-in mechanisms like message passing to adequately model graph structures, LLMs lack such capabilities. To leverage language processing capabilities that LLMs excel at, LLMs rely on preprocessing to linearize graph structures, reducing them from two dimensions to one dimension. This preprocessing, however, significantly increases the prompt length and computational demands~\cite{li2023survey,tan2024musegraph}.
(ii) \textbf{Inter-order Dependency}: The dependency between high-order neighbors and target nodes are complex. As the order of the neighbor increases, the number of neighboring nodes grows geometrically. Capturing these dependencies while maintaining a concise representation for LLMs remains a challenge.

To bridge this gap, we draw inspiration from an intriguing observation: LLMs pre-trained on one language (e.g., English) can achieve exceptional performance on another language (e.g., Chinese) with only a small portion of corpora in the target language~\cite{zhao2023survey}. Building on this insight, we propose \textbf{G}raph- \textbf{D}efined \textbf{L}anguage for \textbf{L}arge \textbf{L}anguage \textbf{M}odel (GDL4LLM), a simple yet effective graph learning with LLMs framework. GDL4LLM enables LLMs to model the text-attributed graph concisely and adequately in a manner analogous to learning a new language. Specifically, we create this new graph language by translating graphs into a graph language corpus and pre-training LLMs to familiarize them with the graph language. Notably, we prove that the pre-training objective enables LLMs to learn graph structural information. We then sample from this corpus to represent the subgraph graph centered around target nodes for fine-tuning on downstream tasks. The corpus captures different orders of structural information using only a few tokens. In all, our contributions can be summarized as follows:

\begin{itemize}[leftmargin=*]
    \item We convert the problem of modeling graph structures for LLMs into a graph language learning problem. We justify this approach by proving that the graph language learning objective enables LLMs to learn graph structural information.

   \item We introduce GDL4LLM, a simple yet effective framework. It generates a graph language corpus from the given graph and pre-trains LLMs on this corpus to understand the graph. The framework then samples from the graph language corpus to represent subgraphs centered around target nodes for fine-tuning on downstream tasks.
    
    \item Through extensive experiments on three real-world datasets, we demonstrate that GDL4LLM outperforms competitive baselines. It surpasses both description-based and textual attribute embedding-based approaches by efficiently modeling different orders of neighbors with LLMs. 
\end{itemize}

\section{Preliminary}
\textbf{Notation.} The text-attributed graph could be represented by a triple \( \mathcal{G} = (\mathcal{V}, \mathcal{E}, \mathcal{X}) \), where \(\mathcal{V} \) denotes the set of nodes \(\{v_{1}, v_{2},\ldots,\) \(v_{|\mathcal{V}|}\}\) with the size \( |\mathcal{V}| \); \( \mathcal{E} \subseteq \mathcal{V} \times \mathcal{V} \) represents the set of edges between nodes with the size \( |\mathcal{E}| \). And the edge set \( \mathcal{E}\) is encoded in the adjacency matrix \( \mathbf{A} \in \mathbb{R}^{|\mathcal{V}| \times |\mathcal{V}|} \), where \( \mathbf{A}_{ij} = 1 \) if there is an edge between nodes \( v_i \) and \( v_j \), \( \mathbf{A}_{ij} = 0 \) otherwise; \( \mathcal{X} = \{\mathbf{x}_1, \mathbf{x}_2, ..., \mathbf{x}_{|\mathcal{V}|}\} \) represents the text attributes associated with each node, where \( \mathbf{x}_i \in \mathcal{X} \) is the node attribute associated with node \( v_i \in \mathcal{V} \). The node attribute \( \mathbf{x}_i \) could be a paper abstract, an item description, or other textual documents. 

\textbf{Node Classification Task.}
Given the graph $\mathcal{G}$, we aim to learn an embedding vector $\mathbf{t}_i$ for each node that captures both the structural information and the semantic information from the node attribute \( \mathbf{x}_i \) to predict the node class label.

For each class, a linear weight vector projects the embedding \(\mathbf{t}_i\) to a class score and the score is passed through a softmax function to obtain the predicted probability \(\hat{y}_i^c\) for class $c$:
\begin{equation}
\hat{y}_i^c = \frac{\exp(\hat{\mathbf{W}}_{h,c}^\top \mathbf{t}_i + b_c)}{\sum_{\hat{c}} \exp(\hat{\mathbf{W}}_{h,\hat{c}}^\top \mathbf{t}_i + b_{\hat{c}})},
\end{equation}

where \(\hat{\mathbf{W}}_{h,c}\) and \(b_c\) are the weight vector and bias for class \(c\) and $\hat{\mathbf{W}}_{h}$ is the unified weight matrix of all vectors. Then we calculate the classification loss $\mathcal{L}_{CE}$ associated with each node based on the predicted probabilities $\hat{y}_i^c$.

We adopt the cross-entropy loss and calculate the loss as follows:
\begin{equation}
\mathcal{L}_{CE} = -\sum_{i=1}^{|\mathcal{V}|} \sum_{c=1}^{C} y_i^c \log(\hat{y}_i^c)
\end{equation}

where \(C\) is the number of classes; \(y_i^c\) is the true class label (1 if node \(i\) belongs to class \(c\), otherwise 0); and \(\hat{y}_i^c\) is the predicted probability of class \(c\).

\textbf{Low Rank Adaption.} To fine-tune LLMs efficiently, we leverage Low-Rank Adaptation (LoRA)~\cite{hu2021lora}. Specifically, given the weight matrix of one layer of LLM $\mathbf{W}_0 \in\mathbb{R}^{d\times d^{\prime}}$, LoRA introduces $\mathbf{W}_0 + \Delta \mathbf{W} =\mathbf{W}_0 + BA$, where $B\in \mathbb{R}^{d\times r}, A\in \mathbb{R}^{r\times d^{\prime}}$, and the rank $r\ll\{d, d^{\prime}\}$, for LLM optimization. $A$ is initialized with Gaussian distribution and $B$ is initialized with zero, therefore, $\Delta \mathbf{W} = BA$ is zero at the beginning of the fine-tuning. During the optimization, the introduced $\Delta \mathbf{W}$ is trained to adapt to the new training data while the original weight $\mathbf{W}_0$ remains frozen.
\begin{figure*}[htbp]
    \centering
    \includegraphics[width=\linewidth]{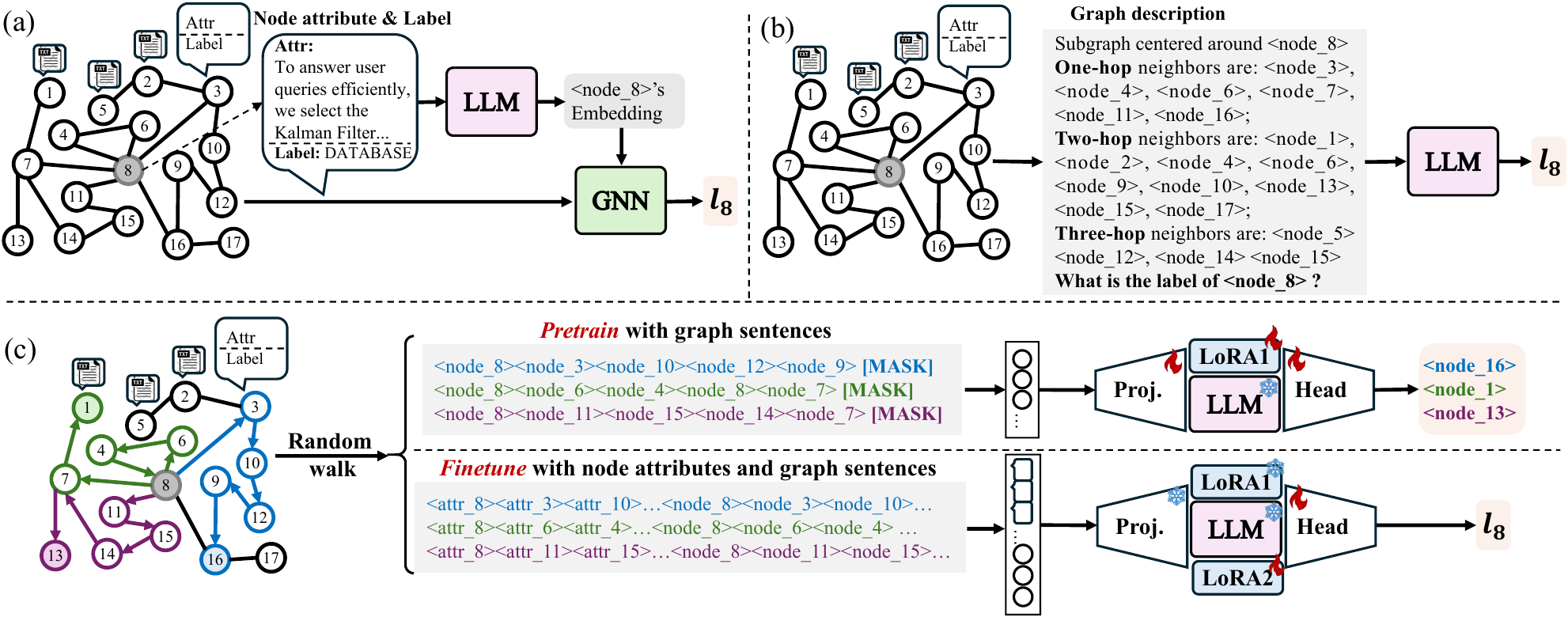}
    \caption{The figure demonstrates a comparison between mainstream methods and GDL4LLM for node-classification task. Figure (a) utilizes LLMs to embed node attributes and leverages GNN to aggregate the embeddings. Figure (b) presents the descriptions of graph structure centered around target nodes. Figure (c) illustrates how LLMs are pre-trained to capture graph structures through graph language learning, and how textual attributes are further integrated to enhance LLMs fine-tuning.}
    \label{fig:main}
\end{figure*}

\section{Methodology}
In this section, we introduce GDL4LLM, a framework for modeling text-attributed graphs through graph language for downstream tasks. As illustrated in Figure~\ref{fig:main}, GDL4LLM operates in two primary stages. In the pre-training stage, we construct a graph language corpus and pre-train LLMs to learn the underlying language patterns. The fine-tuning stage leverages this learned graph language to represent the subgraph centered around the target nodes, adapting the pre-trained LLMs for specific downstream tasks. 

\subsection{Graph Language Pre-training}
We begin by defining the fundamental concepts in graph language, such as graph tokens and graph sentences. We then demonstrate how to derive graph sentences from a given graph to create a graph language corpus and pre-train LLMs on this corpus. And we prove that the pre-training objective encodes graph structural information. 

\subsubsection{Graph Token and Graph Sentence} 

\textbf{Graph Node as Graph Token.} We define the graph token set as equivalent to the node set $\mathcal{V}$. To facilitate the recognition of graph tokens, we extend the LLM tokenizer vocabulary by assigning each node a unique token identifier. For instance, node 1 in the graph is represented as "<node\_1>".

\noindent\textbf{Graph Path as Graph Sentence.} We define a graph sentence as a sequence of edges connecting a series of graph tokens. For example, in Figure~\ref{fig:main}, the first graph sentence in the pre-training stage is <node\_8><node\_3><node\_10><node\_12> <node\_9>. We traverse the graph and translate the graph into a collection of sampled graph sentences formed as a graph language corpus $\mathcal{C}$, i.e., $\mathcal{C} = \{ s_{i}\}, i=\{1, 2, \ldots\}$.

\subsubsection{Graph Corpus Sampling}
\label{sssec:graph corpus sampling}

The key idea in sampling graph sentences is to capture local graph structural information centered around each node. Any graph token can serve as the start token for a graph sentence. To traverse the graph and extend the graph sentence from the start token, we employ random walks to control the traversal process. Given a node $v_i$, the probability of node $v_j$ becoming the next graph token is:

\begin{equation}
P(v_j \mid v_i) = 
\begin{cases}
\frac{\mathbf{A}_{ij}}{\sum_{j \in \mathcal{N}(v_i)} \mathbf{A}_{ij}}, & \text{if } v_j \in \mathcal{N}(v_i), \\
0, & \text{if } e_{ij} \notin \mathcal{E}.
\end{cases}
\end{equation}

where $\mathcal{N}(v_i)$ represents the set of nodes directly connected to $v_i$. By initiating $k$ random walks of length $l$ beginning with each graph token, we can explore both local and high-order graph structural information for each node. Given $|\mathcal{V}|$ nodes, this process samples a corpus of graph language sentences with $k \times |\mathcal{V}|$ size for the entire graph. 

\subsubsection{Pre-training LLMs on the Corpus}

We project out-of-vocabulary graph tokens into LLM-comprehensible embeddings by using LLM to summarize textual attributes $\mathbf{x}_i$. A learnable linear projector $\mathbf{W}_{p} \in \mathbb{R}^{d\times d}$ then aligns these initial embeddings with graph structural information.

For pre-training, we initialize LoRA weight $\Delta \mathbf{W}_1$ and projector $\mathbf{W}_{p}$. During pre-training, $\mathbf{W}_{p}$ maps contextually similar graph tokens to similar embeddings, while $\Delta \mathbf{W}_1$ learns token transition patterns. We maximize the likelihood of predicting the next token in graph sentences and we use one graph sentence $s_i$ as an example:

\begin{equation}
\label{pre-train}
\mathcal{L}_{pre} = -\sum_{q=1}^{l} \log P(s_{i, q} | s_{i, 1:q-1}; \mathbf{W}_{p}, \Delta \mathbf{W}_1, \mathbf{W}_{h}),
\end{equation}

where $\mathbf{W}_{h} \in \mathbb{R}^{d \times |\mathcal{V}|}$ is the trainable LLM head. The next token $s_{i, q}$ is selected based on the highest score from the inner product between $\mathbf{W}_{h}$ and the LLM-generated hidden representation $\mathbf{t}_q$.

\subsubsection{Connection between Pre-training and Graph Learning}
During pre-training, LLMs learn node connections through iterative prediction of graph tokens in the corpus. This learning process is manifested in the inner products between the next token's hidden representation $\mathbf{t}_q$ and its corresponding weight vector $\mathbf{W}_{h,q}$ in $\mathbf{W}_{h}$. The reason is that the pre-training objective maximizes inner products between correct tokens in graph sentences and weight vectors, causing frequently occurring graph sentences to become well-optimized, with their next token representations converging closer to the corresponding weight vectors.

The sampling frequency of graph sentences naturally correlates with node degrees. Specifically, densely connected regions generate lower sampling probabilities for any specific sequence, as they offer more potential graph sentences compared to sparsely connected regions. This relationship between node degree and sampling probability helps encode structural information during pre-training.

\begin{theorem}
\label{theorem1}
For a language model with sufficient capacity to construct the inner product between all $\mathbf{W}_{h,q}$ and $\mathbf{t}_q$ pairs, given node $s_{i,q}$ with degree $d_{q}$, then $\mathbf{W}_{h,q} \cdot \mathbf{t}_q \propto \log \left(\frac{\mathbb{I}_{(s_{i,q-1}, s_{i,q}) \in \mathcal{E}} \cdot \sum_{\mathbf{A}}}{d_q}\right)$.
\end{theorem}

Theorem~\ref{theorem1} demonstrates how node degrees influence the pre-training objective and proof is in the Appendix~\ref{sec:proof}. Higher node degrees result in lower optimal inner product values, while lower degrees lead to higher values, reflecting the certainty of sampled graph sentences. Through process, LLMs effectively learn graph structural information, including node degrees and their connections.

\subsection{Graph Structure-aware Fine-tuning}

We first present the fine-tuning approach for node classification using graph language, followed by a discussion on integrating node textual attributes to enhance both pre-training and fine-tuning stages.

\subsubsection{Absorbing Graph Structure Knowledge}

To preserve pre-trained graph structure knowledge while adapting to new tasks, we merge the pre-trained LoRA weights carefully. Since the pre-trained LoRA parameters contain essential graph structure knowledge, we initialize a new set of LoRA weights $\Delta \mathbf{W}_2$ for fine-tuning rather than modifying existing weights. The LLM parameters are composed as:

\begin{equation}
    \mathbf{W} = \mathbf{W}_0 + \Delta \mathbf{W}_1 + \Delta \mathbf{W}_2.
\end{equation}

We keep $\mathbf{W}_0$ and $\Delta \mathbf{W}_1$ fixed while learning $\Delta \mathbf{W}_2$ during fine-tuning. The head layer weights are reinitialized as \(\hat{\mathbf{W}}_h \in \mathbb{R}^{d \times C}\) for class label adaptation, and we apply LoRA to the fixed projector weights to obtain $\hat{\mathbf{W}}_{p}$ for subsequent node classification optimization.

\subsubsection{Fine-tuning with Graph Language}

For classifying node $v_i$, we sample multiple graph sentences $\{s_1, s_2, \ldots, s_k\}$ of length $l$ starting from $v_i$'s graph token, following sub-subsection~\ref{sssec:graph corpus sampling}. These sentences form a small corpus that captures multi-order structural information around $v_i$. The classification loss for node $v_i$ is:

\begin{equation}
\label{eq:fine-tune}
\mathcal{L}_{CE} = - \log P(y_i | s_{1:k}; \hat{\mathbf{W}}_{p}, \Delta \mathbf{W}_2, \hat{\mathbf{W}}_{h}).
\end{equation}

This small corpus effectively balances local and global structural information through a periodic restart mechanism. While frequent returns to the starting node capture detailed local neighborhood information, appropriate sentence length $l$ enables exploration of higher-order structures.

This two-stage approach offers key advantages: (i) LLMs learn the graph structures sufficiently through language-like pre-training, eliminating the need for verbose descriptions, and (ii) concise representation of different orders of graph structures through sampled graph language corpus during fine-tuning, leading to efficient node classification.

\subsubsection{Fine-tuning with Textual Node Attributes}
\label{sssec:fntext}

To leverage both structural and semantic information, we combine graph language corpus with textual node attributes. While graph language captures structural patterns, incorporating textual attributes enriches the prompt with semantic information. We create a composite document for each node by traversing the graph language corpus and appending each visited node's textual attributes. This document accompanies the graph language in both pre-training and fine-tuning stages, enabling LLMs to comprehend node connections and generate accurate classifications through their natural language understanding capabilities.

\section{Experiments}
We conduct experiments across three datasets to evaluate GDL4LLM's performance and analyze its behavior. We aim to address following four research questions: \textbf{RQ1}: How does GDL4LLM compare to state-of-the-art text-attributed frameworks in node classification tasks? \textbf{RQ2}: How do the pre-training stage and textual attributes in the prompt contribute to the overall model performance? \textbf{RQ3}: How do key hyperparameters, including graph sentence length and choice of LLM backbone, affect model performance? \textbf{RQ4}: How efficiently do GDL4LLM capture different orders of graph structures?

\subsection{Dataset}
We evaluate the performance of GDL4LLM using three datasets: ACM, Wikipedia, and Amazon. These datasets have been manually created from the raw corpus along with their respective descriptions. The statistics for these three datasets are presented in Table~\ref{tab:dataset}. Wiki. The raw data consists of text in all languages from Wikipedia
articles. ACM~\cite{tang2008arnetminer}. The ACM dataset contains papers published in flagship conferences, such as KDD and SIGMOD. Amazon. The Amazon dataset comprises product 
 metadata from the famous e-commerce website~\cite{ni2019justifying}.
\begin{table}
  \caption{Statistics of datasets in our experiment.}
  \label{tab:dataset}
  \centering
  \setlength{\tabcolsep}{1mm}
  \begin{tabular}{ccccc}
    \toprule
    Datasets & \#nodes & \#edges  & \#classes\\
    \midrule
    ACM &$48$,$579$ & $193$,$034$ & 9\\
    Wiki & $36$,$501$ & $1$,$190$,$369$ & 10\\
    Amazon & $50$,$000$ & $632$,$802$ & 7\\
  \bottomrule
\end{tabular}
\end{table}

\subsection{Baselines}
We incorporate the following state-of-the-art baselines in our main comparison:  GCN~\cite{kipf2016semi}, GraphSAGE~\cite{hamilton2017inductive}, GAT~\cite{velivckovic2017graph},  MPAD~\cite{nikolentzos2020message}, GLEM~\cite{zhao2022learning}, GraphFormers~\cite{yang2021graphformers}, LLAGA~\cite{chen2024llaga}, InstructGLM~\cite{ye2024language}, and GraphAdapter~\cite{huang2024can} frameworks. Besides these baselines, we also incorporate several variants of GDL4LLM for ablation study: (i) GDL4LLM w/o pre-train: removes the pre-training stage of GDL4LLM and preserves the fine-tuning stage; (ii) GDL4LLM w/ attr: incorporates the textual attributes within the prompt of GDL4LLM; (iii) GDL4LLM w/ attr w/o pre-train: removes the pre-training stage of GDL4LLM w/ attr.

\begin{table*}[h]
\small
\centering
\caption{Node classification performance comparison among baselines w.r.t. micro classification accuracy across three datasets.}
\label{tab:micro_unified}
\setlength{\tabcolsep}{4mm}
\begin{tabular}{llcccccccc}  
\toprule
\multirow{2}{*}{\textbf{NLP Models}} 
& \multirow{2}{*}{\textbf{GNNs}} 
& \multicolumn{2}{c}{\textbf{ACM}} 
& \multicolumn{2}{c}{\textbf{Wiki}} 
& \multicolumn{2}{c}{\textbf{Amazon}} \\ 
\cmidrule(lr){3-4} \cmidrule(lr){5-6} \cmidrule(lr){7-8}
& & Val. & Test & Val. & Test & Val. & Test \\
\midrule

\multicolumn{8}{c}{\textbf{Fine-tuned LMs + GNNs}} \\
\cmidrule{1-8}
\multirow{4}{*}{Bert} & - & 74.4 & 73.2 & 69.5 & 68.8 & 86.2 & 87.0 \\
    & GCN & 77.6 & 77.1 & 69.4 & 68.4 & 92.3 & 92.8 \\
    & GAT & 77.9 & 78.0 & 70.5 & 69.8 & 92.5 & 92.4 \\
    & GraphSAGE & 77.3 & 76.8 & 73.1 & \underline{72.7} & 92.0 & 92.3 \\
\midrule
\multirow{4}{*}{Roberta} & - & 78.1 & 76.6 & 67.8 & 68.1 & 84.9 & 85.9 \\
    & GCN & 80.1 & 79.4 & 68.5 & 68.0 & 92.3 & 92.5 \\
    & GAT & 79.7 & 78.9 & 70.1 & 71.0 & 92.5 & 92.4 \\
    & GraphSAGE & 78.5 & 78.3 & 72.7 & 72.1 & 92.2 & 92.1 \\
& GraphSAGE    & 80.9 & 79.5 & \underline{73.2} & 70.4 & 94.3 & 94.1 \\

\cmidrule{1-8}
\multicolumn{8}{c}{\textbf{Specialized Frameworks for Text-Attributed Graphs}} \\
\cmidrule{1-8}
\multicolumn{2}{c}{MPAD}          & 80.1 & 78.9 & 68.8 & 68.0 & 93.1 & 92.8 \\
\multicolumn{2}{c}{GLEM} & \underline{81.4} & \underline{79.8} & 72.6 & 71.2 & 92.5 & 93.3 \\
\multicolumn{2}{c}{GraphFormers}  & 75.3 & 75.1 & 66.8 & 67.5 & 85.6 & 86.4 \\
\multicolumn{2}{c}{LLAGA}         & 77.2 & 77.5 & 71.7 & 72.0 & 90.1 & 90.8 \\
\multicolumn{2}{c}{InstructGLM}   & 75.4 & 74.5 & 72.2 & 70.6 & \underline{94.3} & \underline{94.2} \\
\cmidrule{1-8}
\multicolumn{2}{c}{\textbf{GDL4LLM}} & \textbf{81.9} & \textbf{81.4} & \textbf{74.3} & \textbf{73.2} & \textbf{94.6} & \textbf{94.6} \\

\cmidrule{1-8}
\multicolumn{8}{c}{\textbf{Fine-tuned Large Language Models +/- GNNs}} \\
\cmidrule{1-8}
GraphAdapter & - & 80.8 & 80.4 & 71.9 & 71.7 & \underline{94.1} & \underline{93.4} \\
Llama3-8b    & - & 80.7 & 80.6 & 71.9 & 71.2 & 92.0 & 91.6 \\
Llama3-8b    & GraphSAGE & \underline{82.0} & \underline{81.3} & \underline{72.8} & \underline{73.0} & 93.1 & 92.8 \\
\cmidrule{1-8}
\multicolumn{2}{c}{\textbf{GDL4LLM w/ attr}} & \textbf{83.9} & \textbf{82.8} & \textbf{74.0} & \textbf{73.4} & \textbf{95.8} & \textbf{95.5} \\

\bottomrule
\end{tabular}

\end{table*}

\subsection{Experimental Setup}
 For node classification tasks, we evaluate performance using micro classification accuracy. In the main comparison, we use Llama2-7B~\citep{touvron2023llama} as the backbone and run all models ten times to report the average results. We conduct experiments on a server with eight 80 GB NVIDIA A100 GPUs. For all baselines, we use publicly released codes to ensure fairness. We implement our framework with Pytorch and transformers packages. To select the hyperparameters, we use grid search strategy. Specifically, the length of the graph sentences \(l\), and the numbers of graph sentences \(k\), are searched using grids \(\{2, 3, 4, 5\}\) and \(\{2, 4, 6, 8, 10\}\), respectively, based on the model performance on the validation set. The LoRA dropout rate, rank, and \(\alpha\) are set to 0.2, 8, and 16, respectively. Weight decay is set to 1e-2 to prevent the overfitting. We implement the proposed framework in PyTorch, with a learning rate of 1e-4 and a batch size of 32. Our approach uses the Adam optimizer and includes early stopping based on validation set micro accuracy. In the main comparison, we use Llama2-7B~\citep{touvron2023llama} as the backbone and employ Llama3-8B for the different backbone experiments. 
 
\subsection{Main Comparison (RQ1)}
Table~\ref{tab:micro_unified} presents the micro classification accuracy comparing GDL4LLM with baseline methods for node classification tasks. We evaluate against three categories of baselines and have the following observations:

(i) \textbf{LM-GNN category}: These methods combine fine-tuned language models with GNNs for classification. While achieving satisfactory results, their performance primarily stems from a combination of GNNs' structural modeling capabilities rather than the sole limited-capacity language models. Their effectiveness is constrained by both LM capacity and insufficient structural information in textual attributes to get better embeddings.

2) \textbf{Description-based category}: These approaches attempt to model graph structure through textual descriptions. However, due to lengthy prompts, they typically only model first-order or second-order structural information. In contrast, GDL4LLM efficiently represents complex graph structures using concise graph tokens, enabling exploration of higher-order structural patterns and outperforming baselines in this category.

3) \textbf{LLM-GNN category}: These methods leverage LLMs to generate high-quality textual attribute embeddings compared to the first group and improve performance. However, they lack deep integration between LLMs and GNN aggregation, which limits LLMs' capability in modeling high-order neighbor relationships. GDL4LLM addresses this limitation by enabling LLMs to comprehend graph structure as a language while utilizing their natural language understanding for textual attributes, resulting in superior performance.

\subsection{Ablation Study (RQ2)}

\begin{figure}[h]
    \centering
    \includegraphics[width=\columnwidth]{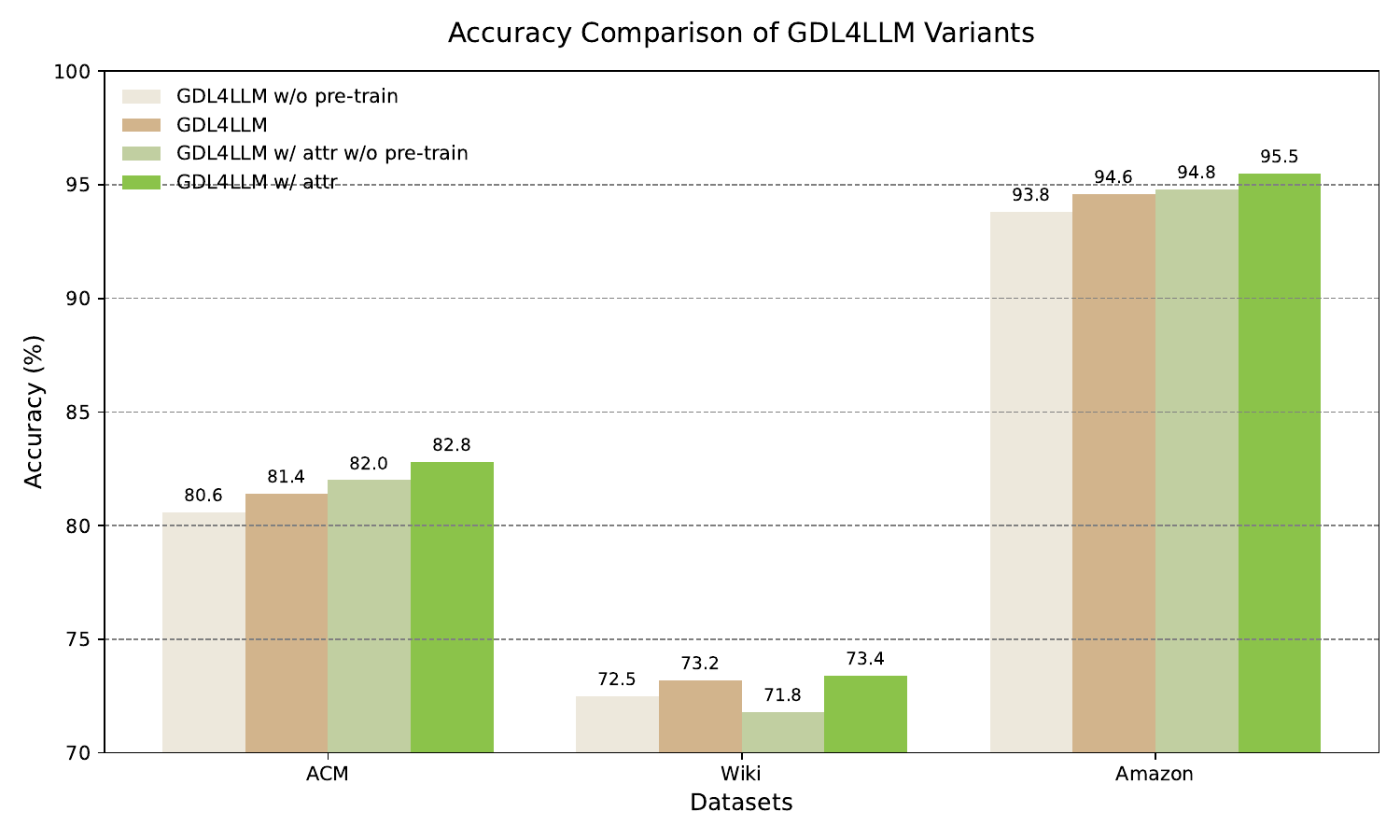}
    \caption{Accuracy comparison of different GDL4LLM variants on the test set across three datasets.}
    \label{fig:tab-3}
\end{figure}

We conduct ablation study to examine how different components affect GDL4LLM's performance, focusing on the pre-training stage and the incorporation of textual attributes. Results are presented in Figure~\ref{fig:tab-3}.

Our analysis reveals that pre-training contributes notably to model performance, especially for GDL4LLM w/ attr. This gain mirrors the benefits of pre-training, as LLMs develop linguistic competence through initial pre-training. The inclusion of textual attributes in prompts alongside graph language corpus further enhances model performance. This improvement demonstrates how GDL4LLM effectively leverages LLMs' natural language understanding capabilities to process node attributes. The combination of pre-training and textual attribute integration creates a synergistic effect, where structural understanding from pre-training complements semantic comprehension from textual attributes.

These findings suggest that both components play distinct yet complementary roles: pre-training establishes foundational graph structure understanding, while textual attribute integration enables richer node representations through natural language processing capabilities.

\begin{table*}[h]
    \small
    \centering
    \caption{Comparison w.r.t. the number of used tokens and the order of graph structure modeled (Token/(order)).}
    \label{tab:efficiency1}
\setlength{\tabcolsep}{4mm}
\begin{tabular}{llcccccccc}  
    \toprule
    \multirow{2}{*}{\textbf{LLMs}} 
    & \multirow{2}{*}{\textbf{-}}
    & \multicolumn{2}{c}{\textbf{ACM}} 
    & \multicolumn{2}{c}{\textbf{Wiki}} 
    & \multicolumn{2}{c}{\textbf{Amazon}} \\ 
    \cmidrule(lr){3-4} \cmidrule(lr){5-6} \cmidrule(lr){7-8}
    & & \textbf{Val.} & \textbf{Test} & \textbf{Val.} & \textbf{Test} & \textbf{Val.} & \textbf{Test} \\ 
    \midrule
    \multicolumn{2}{l}{\textbf{InstructGLM}} 
    & 146(1) & 149(1) & 1024(2) & 319(1) & 532(2) & 538(2) \\ 
    \multicolumn{2}{l}{\textbf{LLAGA-HO}} 
    & 155(4) & 155(4) & 130(4) & 130(4) & 140(4) & 140(4) \\ 
    \multicolumn{2}{l}{\textbf{GDL4LLM}} 
    & 54(4) & 54(4) & 72(4) & 72(4) & 72(4) & 72(4) \\ 
    \cmidrule{1-8}
    \multicolumn{2}{l}{\textbf{Reduction (\%)}} 
    & \textbf{63.01} & \textbf{63.01} 
    & \textbf{44.62} & \textbf{44.62} 
    & \textbf{48.57} & \textbf{48.57} \\ 
    \bottomrule
    \end{tabular}
    \vspace{-2mm}
\end{table*}

\begin{table*}[h]
    \small
    \centering
    \caption{Training and test time comparison across LLMs for three datasets. Times are presented in `hh:mm:ss` format.}
    \label{tab:efficiency2}
\setlength{\tabcolsep}{4mm}
\begin{tabular}{llcccccccc}  
    \toprule
    \multirow{2}{*}{\textbf{LLMs}} 
    & \multirow{2}{*}{\textbf{-}}
    & \multicolumn{2}{c}{\textbf{ACM}} 
    & \multicolumn{2}{c}{\textbf{Wiki}} 
    & \multicolumn{2}{c}{\textbf{Amazon}} \\ 
    \cmidrule(lr){3-4} \cmidrule(lr){5-6} \cmidrule(lr){7-8}
    & & \textbf{Train} & \textbf{Test} & \textbf{Train} & \textbf{Test} & \textbf{Train} & \textbf{Test} \\ 
    \midrule
    \multicolumn{2}{l}{\textbf{InstructGLM}} 
    & 5:59:04 & 0:06:41 & 4:30:16 & 0:05:01 & 6:07:02 & 0:07:08 \\ 
    \multicolumn{2}{l}{\textbf{LLAGA-HO}} 
    & \textbf{0:47:26} & 0:04:41 & 0:35:18 & 0:01:57 & 0:49:29 & 0:04:04 \\ 
    \multicolumn{2}{l}{\textbf{GDL4LLM}} 
    & 1:05:23 & \textbf{0:02:06} & \textbf{0:34:03} & \textbf{0:01:17} & \textbf{0:32:15} & \textbf{0:01:45} \\ 
    \bottomrule
    \end{tabular}
    \vspace{-2mm}
\end{table*}

\subsection{Backbones \& Hyperparameters (RQ3)}

\begin{figure}[h]
    \centering
    \includegraphics[width=\columnwidth]{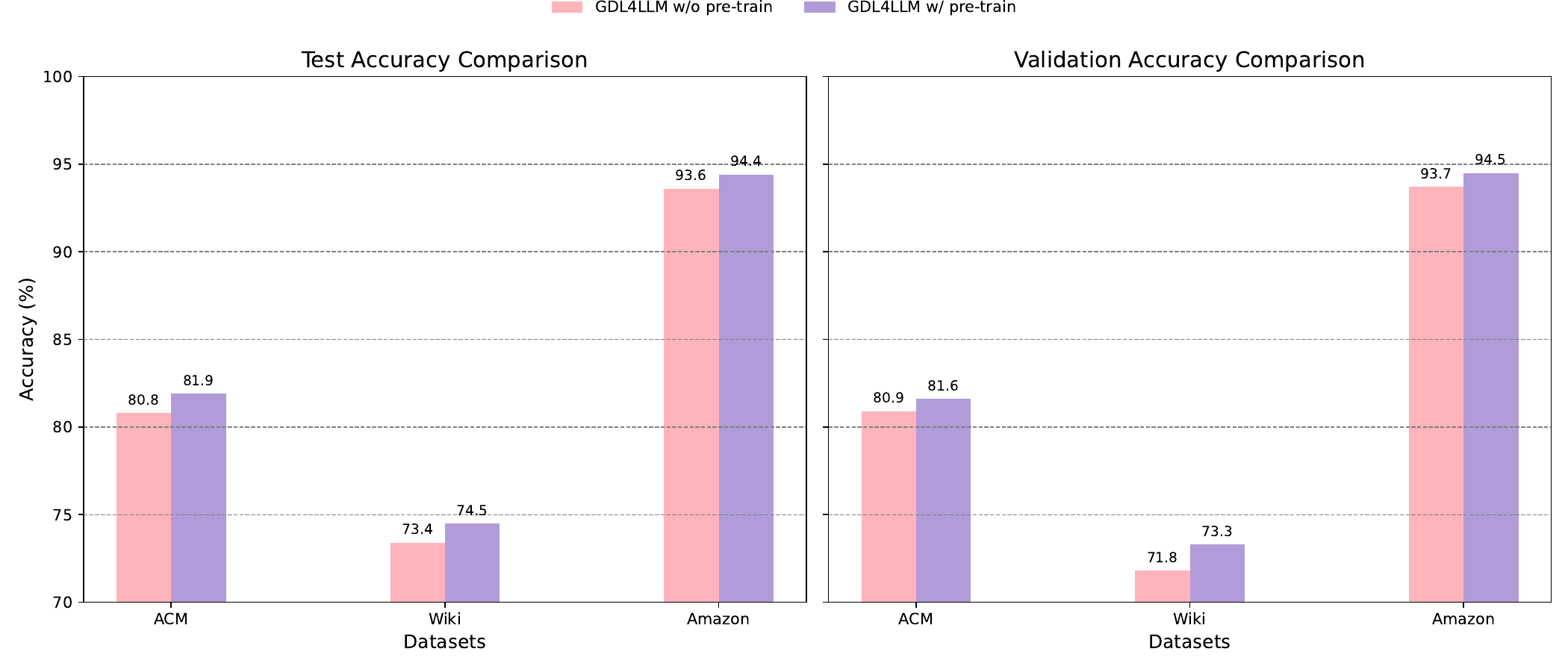}
    \caption{Performance comparison between Llama-2 and Llama-3 backbones on validation and test sets across three datasets.}
    \label{fig:tab-4}
\end{figure}

\noindent\textbf{Backbones.} We evaluate GDL4LLM's performance across different LLM backbones, comparing Llama-2 with Llama-3, as shown in Figure~\ref{fig:tab-4}. Results indicate that Llama-3 achieves better performance, likely due to its improved architecture and enhanced training data quality. Ablation studies on Llama-3 confirm the benefits of our pre-training objective across different LLM architectures.

\begin{figure}[htbp]
    \centering
    \includegraphics[width=0.8\linewidth,trim=0.8cm 0cm 0.8cm 0.8cm, clip]{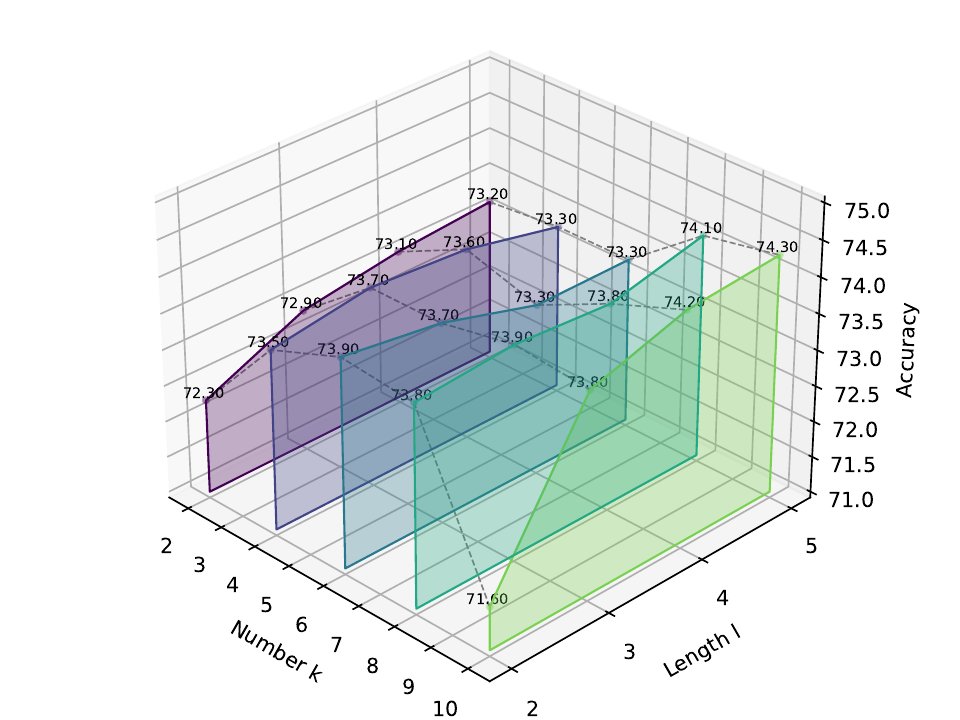}
    \caption{Visualizations of the impact of graph sentence length $l$ and graph sentence length $k$ on performance.}
    \label{fig:tab-6}
\end{figure}

\noindent\textbf{Hyperparameters.} We examine two critical hyperparameters: the length of sampled graph sentences $l$ and the number of sampled sentences $k$. Figure~\ref{fig:tab-6} shows optimal performance at $l=5$ and $k=10$, and the performance gain is marginal when approaching this value. These results demonstrate our framework's effectiveness in modeling high-order structural information, such as inter-order dependencies. For instance, a length of $5$ captures fourth-order structural information, whereas GNNs, often converging in about two layers, typically capture only second-order information~\cite{chen2024text,song2024pure}.

These findings align with our hypothesis that GDL4LLM can achieve strong performance in modeling inter-order dependency and can achieve generalization in different backbones.

\subsection{Efficiency Analysis (RQ4)}

We analyze GDL4LLM's efficiency in modeling high-order structural information by comparing token usage and running time against description-based frameworks. Table~\ref{tab:efficiency1} demonstrates that GDL4LLM achieves significant token reduction compared to InstructGLM and LLAGA-HO. While LLAGA-HO employs GNNs for neighbor summarization, its reliance on natural language task descriptions leads to verbose prompts. In contrast, GDL4LLM's succinct graph language prompts eliminate this verbosity while enabling exploration of higher-order graph information.

The computational advantages of GDL4LLM are evident in Table~\ref{tab:efficiency2}, showing faster training and inference across most datasets. On the ACM dataset, GDL4LLM reduces inference time to 2:06, compared to InstructGLM (6:41) and LLAGA-HO (4:41). These results demonstrate GDL4LLM's ability to significantly reduce token usage and running time with effective structure modeling.

\section{Related Work}
\subsection{GNNs for Text-attributed Graph}
In traditional pipelines for text-attributed graph analysis, natural language processing techniques are first employed to extract features from textual data, which are then utilized in graph neural networks (GNNs) for graph propagation. Common NLP methods include Bag of Words \cite{zhang2010understanding}, fixed embeddings like Word2Vec and GloVe\cite{pennington2014glove}, as well as the use of pre-trained models such as BERT and fine-tuned variations \cite{liu2019roberta,gao2021simcse}. These approaches establish the foundational representation of textual information within the graph structure. However, more recent advancements have led to the development of tailored graph learning methods specifically designed for text-attributed graphs. Notable examples include Graphformers \cite{yang2021graphformers} and MPAD \cite{nikolentzos2020message}, which represent word-adjacent relationships as graphs, offering an alternative perspective to text-attributed graph representations. GLEM \cite{zhao2022learning} integrates graph structure and language learning using a variational Expectation-Maximization framework. Other tailored approaches \cite{chien2021node,duan2023simteg} aim to improve the flexibility and efficacy of graph-based analyses from complex and structured data. 

\subsection{LLMs for Text-attributed Graph}
Deep learning on graphs has also achieved great success, ranging from recommendation system ~\cite{zhou2023adaptive,zhou2024enhancing}, representation learning~\cite{huang2022going} and text-attributed networks~\cite{sun2022does}.  Recent advancements in LLMs~\cite{li_Survey_2024,jin2023patton} have introduced novel opportunities for tasks involving text-attributed graphs. By properly designing prompts, researchers can instruct LLMs to generate contextually enriched node and edge representations~\cite{zhou2024quest}, facilitating various downstream tasks such as node classification \cite{he_Harnessing_2024,seo_Unleashing_2024,zhu_Efficient_2024}. Efficiently training such LLMs has become a priority. Recent techniques such as LoRA (Low-Rank Adaptation) \cite{hu2021lora,dettmers2024qlora} enable fine-tuning of pre-trained models by injecting trainable adapters, significantly reducing memory usage and computational costs. Besides, prefix-tuning \cite{li2021prefix,lester2021power} optimizes model performance by learning a fixed set of prefixes that condition the model's responses without modifying the entire architecture. Fine-tuning LLMs for graphs often requires costly labeled data. To address this, SFGL~\cite{luscale} leverages the scale-free property of real-world graphs and a graph-based pseudo-labeler to enhance LLM fine-tuning without heavy supervision. LLM-GNN~\cite{chenlabel} uses LLMs to generate confidence-aware pseudo-labels and selects informative nodes, enabling effective training without human annotations. Meanwhile, Latent Graph Inference with Limited Supervision~\cite{lu2023latent} improves generalization by restoring lost connections in sparse graphs and rebalancing learning signals. Together, these methods reduce the need for labeled data by combining structural priors, pseudo-labeling, and efficient supervision. We incorporate these ideas into model design and achieve a higher accuracy without sacrificing efficiency.

\section{Conclusion}
In this paper, we identify two key limitations of using natural language alone to model text-attributed graphs with LLMs: (i) Graph descriptions become excessively verbose, particularly when modeling high-order neighbors, and (ii) textual attributes often lack sufficient structural information, limiting LLMs' ability to capture high-order graph structures. We address two main challenges: (i) the absence of a built-in mechanism in LLMs, akin to message passing in GNNs, and (ii) the inter-order dependencies between high-order neighbors and target nodes. To overcome these challenges, we propose the GDL4LLM framework, which enables LLMs to model text-attributed graphs as if learning a new language. GDL4LLM consists of two main stages: (i) collecting a graph language corpus and pre-training LLMs on this corpus to adequately understand the graph, and (ii) fine-tuning LLMs with this corpus to concisely represent subgraph centered around target nodes. This approach allows LLMs to effectively capture graph structure of various orders. Extensive experiments demonstrate the efficiency and effectiveness of GDL4LLM in modeling text-attributed graphs for downstream tasks. Our future work will explore more efficient graph learning methods with graph language.

\section{Proof of Theorem~\ref{theorem1}}
\label{sec:proof}
Assume that all possible token weight vectors \(\mathbf{W}_{\text{h,q}}\) form a weight vector set \(\mathcal{W}\), and all hidden representations of the next possible graph token \(\mathbf{t}_q\) form a context set \(\mathcal{T}\). We use \(\#(\cdot, \cdot)\) to represent the number of co-occurrences of two quantities. If the language model has sufficient capacity to construct the inner product between any pair \(\mathbf{W}_{\text{h,q}}\) and \(\mathbf{t}_q\), the optimization of each pair's inner product does not affect the others. Then we replace the cross entropy loss in the pre-training process with the binary cross entropy loss. The overall pre-training loss is calculated as:

\begin{align}
\mathcal{L} = & \sum_{\mathbf{w}\in \mathcal{W}} \sum_{\mathbf{t}\in \mathcal{T}} \#(w,t) 
\left(\log \left(\frac{\exp(\mathbf{w} \cdot \mathbf{t})}{1 + \exp(\mathbf{w} \cdot \mathbf{t})}\right)\right) \nonumber \\
& + \sum_{\mathbf{w}\in \mathcal{W}} \sum_{\mathbf{t}\in \mathcal{T}} \#(w,t) \cdot |\mathcal{W}| \cdot \mathbb{E}_{t \sim P_t} \nonumber \\
& \quad \left[\log \left(\frac{1}{1 + \exp(\mathbf{w} \cdot \mathbf{t})}\right)\right].
\end{align}

The pre-training loss with respect to a specific pair is calculated as:

\begin{align}
\mathcal{L}_{(\mathbf{w},\mathbf{t})} = & \#(\mathbf{w},\mathbf{t}) 
\log \left(\frac{\exp(\mathbf{w} \cdot \mathbf{t})}{1 + \exp(\mathbf{w} \cdot \mathbf{t})}\right) \nonumber \\
& + |\mathcal{W}| \cdot \#(\mathbf{w}) \cdot k 
\log \left(\frac{1}{1 + \exp(\mathbf{w} \cdot \mathbf{t})}\right),
\end{align}
where $k$ is a constant factor. We calculate the gradient of the above term and set it to zero, obtaining the following terms:

\begin{align}
e^{2x} - \left(\frac{\#(\mathbf{w}, \mathbf{t})}{|\mathcal{W}| \cdot \#(\mathbf{w}) \cdot d_q} - 1\right) e^x 
= \frac{\#(\mathbf{w},\mathbf{t})}{|\mathcal{W}| \cdot \#(\mathbf{w}) \cdot k}.
\end{align}

The inner product \(\mathbf{w} \cdot \mathbf{t}\), i.e., \(x\), satisfies the following equation:

\begin{align}
\mathbf{w} \cdot \mathbf{t} \propto 
\log \left(\frac{\mathbb{I}_{(s_{i,q-1}, s_{i,q}) \in \mathcal{E}} \cdot \sum_{\mathbf{A}}}{d_q d_{q-1}}\right).
\end{align}

\section*{Acknowledgement}
The work described in this paper was fully supported by a grant from the Innovation and Technology Commission of the Hong Kong Special Administrative Region, China (Project No. GHP/391/22).

\section*{Limitations}
While our approach of treating graph learning as language learning for LLMs demonstrates significant improvements in efficiency compared to description-based methods, certain limitations remain. The primary constraint is the memory consumption inherent to LLMs, which may restrict deployment in resource-constrained environments. To address this limitation, we use LoRA to reduce memory consumption and load standard quantized LLMs instead of full precision models for both our models and all baselines. And future work will explore the possibility of further memory consumption through LLM layer reduction while maintaining model performance.

\section*{Ethics Statement}
Our study utilizes three datasets: the widely-used Amazon dataset, and two custom datasets (Wiki and ACM) built from public sources. For the custom datasets, we protect user privacy through thorough anonymization of consumer information. All baseline implementations are publicly available through open-source repositories. In the preparation of this work, AI-based tools are utilized exclusively for polishing purposes, such as refining the clarity and grammar of the text. These tools are not employed for generating ideas, or conducting research, ensuring that all scientific contributions and implementations are entirely original. Our research strictly adheres to the ACM Code of Ethics\footnote{https://www.acm.org/code-of-ethics}, particularly regarding data privacy, transparency, and responsible computing practices.

\bibliography{main}

\begin{thebibliography}{49}
\providecommand{\natexlab}[1]{#1}

\bibitem[{Achiam et~al.(2023)Achiam, Adler, Agarwal, Ahmad, Akkaya, Aleman, Almeida, Altenschmidt, Altman, Anadkat et~al.}]{achiam2023gpt}
Josh Achiam, Steven Adler, Sandhini Agarwal, Lama Ahmad, Ilge Akkaya, Florencia~Leoni Aleman, Diogo Almeida, Janko Altenschmidt, Sam Altman, Shyamal Anadkat, and 1 others. 2023.
\newblock Gpt-4 technical report.
\newblock \emph{arXiv preprint arXiv:2303.08774}.

\bibitem[{Chen et~al.(2024{\natexlab{a}})Chen, Zhao, Jaiswal, Shah, and Wang}]{chen2024llaga}
Runjin Chen, Tong Zhao, Ajay Jaiswal, Neil Shah, and Zhangyang Wang. 2024{\natexlab{a}}.
\newblock Llaga: Large language and graph assistant.
\newblock \emph{arXiv preprint arXiv:2402.08170}.

\bibitem[{Chen et~al.(2024{\natexlab{b}})Chen, Mao, Li, Jin, Wen, Wei, Wang, Yin, Fan, Liu et~al.}]{chen2024exploring}
Zhikai Chen, Haitao Mao, Hang Li, Wei Jin, Hongzhi Wen, Xiaochi Wei, Shuaiqiang Wang, Dawei Yin, Wenqi Fan, Hui Liu, and 1 others. 2024{\natexlab{b}}.
\newblock Exploring the potential of large language models (llms) in learning on graphs.
\newblock \emph{ACM SIGKDD Explorations Newsletter}, 25(2):42--61.

\bibitem[{Chen et~al.(2024{\natexlab{c}})Chen, Mao, Liu, Song, Li, Jin, Fatemi, Tsitsulin, Perozzi, Liu et~al.}]{chen2024text}
Zhikai Chen, Haitao Mao, Jingzhe Liu, Yu~Song, Bingheng Li, Wei Jin, Bahare Fatemi, Anton Tsitsulin, Bryan Perozzi, Hui Liu, and 1 others. 2024{\natexlab{c}}.
\newblock Text-space graph foundation models: Comprehensive benchmarks and new insights.
\newblock \emph{arXiv preprint arXiv:2406.10727}.

\bibitem[{Chen et~al.()Chen, Mao, Wen, Han, Jin, Zhang, Liu, and Tang}]{chenlabel}
Zhikai Chen, Haitao Mao, Hongzhi Wen, Haoyu Han, Wei Jin, Haiyang Zhang, Hui Liu, and Jiliang Tang.
\newblock Label-free node classification on graphs with large language models (llms).
\newblock In \emph{The Twelfth International Conference on Learning Representations}.

\bibitem[{Chien et~al.(2021)Chien, Chang, Hsieh, Yu, Zhang, Milenkovic, and Dhillon}]{chien2021node}
Eli Chien, Wei-Cheng Chang, Cho-Jui Hsieh, Hsiang-Fu Yu, Jiong Zhang, Olgica Milenkovic, and Inderjit~S Dhillon. 2021.
\newblock Node feature extraction by self-supervised multi-scale neighborhood prediction.
\newblock \emph{arXiv preprint arXiv:2111.00064}.

\bibitem[{Cook and Holder(2006)}]{cook2006mining}
Diane~J Cook and Lawrence~B Holder. 2006.
\newblock \emph{Mining graph data}.
\newblock John Wiley \& Sons.

\bibitem[{Dettmers et~al.(2024)Dettmers, Pagnoni, Holtzman, and Zettlemoyer}]{dettmers2024qlora}
Tim Dettmers, Artidoro Pagnoni, Ari Holtzman, and Luke Zettlemoyer. 2024.
\newblock Qlora: Efficient finetuning of quantized llms.
\newblock \emph{Advances in Neural Information Processing Systems}, 36.

\bibitem[{Duan et~al.(2023)Duan, Liu, Chua, Yan, Ooi, Xie, and He}]{duan2023simteg}
Keyu Duan, Qian Liu, Tat-Seng Chua, Shuicheng Yan, Wei~Tsang Ooi, Qizhe Xie, and Junxian He. 2023.
\newblock Simteg: A frustratingly simple approach improves textual graph learning.
\newblock \emph{arXiv preprint arXiv:2308.02565}.

\bibitem[{Fatemi et~al.(2024)Fatemi, Halcrow, and Perozzi}]{fatemitalk}
Bahare Fatemi, Jonathan Halcrow, and Bryan Perozzi. 2024.
\newblock Talk like a graph: Encoding graphs for large language models.
\newblock In \emph{The Twelfth International Conference on Learning Representations}.

\bibitem[{Gao et~al.(2021)Gao, Yao, and Chen}]{gao2021simcse}
Tianyu Gao, Xingcheng Yao, and Danqi Chen. 2021.
\newblock Simcse: Simple contrastive learning of sentence embeddings.
\newblock \emph{arXiv preprint arXiv:2104.08821}.

\bibitem[{Guan et~al.(2024)Guan, Zhao, Wu, He, and Fan}]{guan2024langtopo}
Zhong Guan, Hongke Zhao, Likang Wu, Ming He, and Jianpin Fan. 2024.
\newblock Langtopo: Aligning language descriptions of graphs with tokenized topological modeling.
\newblock \emph{arXiv preprint arXiv:2406.13250}.

\bibitem[{Guo et~al.(2024)Guo, Xia, Yu, Wang, Yang, Wei, Pang, Chua, and Huang}]{guo2024graphedit}
Zirui Guo, Lianghao Xia, Yanhua Yu, Yuling Wang, Zixuan Yang, Wei Wei, Liang Pang, Tat-Seng Chua, and Chao Huang. 2024.
\newblock Graphedit: Large language models for graph structure learning.
\newblock \emph{arXiv preprint arXiv:2402.15183}.

\bibitem[{Hamilton et~al.(2017)Hamilton, Ying, and Leskovec}]{hamilton2017inductive}
Will Hamilton, Zhitao Ying, and Jure Leskovec. 2017.
\newblock Inductive representation learning on large graphs.
\newblock \emph{Advances in neural information processing systems}, 30.

\bibitem[{He et~al.(2024)He, Bresson, Laurent, Perold, LeCun, and Hooi}]{he_Harnessing_2024}
Xiaoxin He, Xavier Bresson, Thomas Laurent, Adam Perold, Yann LeCun, and Bryan Hooi. 2024.
\newblock \href {https://arxiv.org/abs/2305.19523} {Harnessing {{Explanations}}: {{LLM-to-LM Interpreter}} for {{Enhanced Text-Attributed Graph Representation Learning}}}.
\newblock \emph{Preprint}, arXiv:2305.19523.

\bibitem[{Hong et~al.(2024)Hong, Yuan, Zhang, Chen, Dong, Huang, and Huang}]{hong2024next}
Zijin Hong, Zheng Yuan, Qinggang Zhang, Hao Chen, Junnan Dong, Feiran Huang, and Xiao Huang. 2024.
\newblock Next-generation database interfaces: A survey of llm-based text-to-sql.
\newblock \emph{arXiv preprint arXiv:2406.08426}.

\bibitem[{Hu et~al.(2021)Hu, Shen, Wallis, Allen-Zhu, Li, Wang, Wang, and Chen}]{hu2021lora}
Edward~J Hu, Yelong Shen, Phillip Wallis, Zeyuan Allen-Zhu, Yuanzhi Li, Shean Wang, Lu~Wang, and Weizhu Chen. 2021.
\newblock Lora: Low-rank adaptation of large language models.
\newblock \emph{arXiv preprint arXiv:2106.09685}.

\bibitem[{Huang et~al.(2019)Huang, Song, Li, and Hu}]{huang2019graph}
Xiao Huang, Qingquan Song, Yuening Li, and Xia Hu. 2019.
\newblock Graph recurrent networks with attributed random walks.
\newblock In \emph{Proceedings of the 25th ACM SIGKDD International Conference on Knowledge Discovery \& Data Mining}, pages 732--740.

\bibitem[{Huang et~al.(2024)Huang, Han, Yang, Bao, Tao, Chai, and Zhu}]{huang2024can}
Xuanwen Huang, Kaiqiao Han, Yang Yang, Dezheng Bao, Quanjin Tao, Ziwei Chai, and Qi~Zhu. 2024.
\newblock Can gnn be good adapter for llms?
\newblock In \emph{Proceedings of the ACM on Web Conference 2024}, pages 893--904.

\bibitem[{Huang et~al.(2022)Huang, Wang, Li, and He}]{huang2022going}
Zhongyu Huang, Yingheng Wang, Chaozhuo Li, and Huiguang He. 2022.
\newblock Going deeper into permutation-sensitive graph neural networks.
\newblock In \emph{International Conference on Machine Learning}, pages 9377--9409. PMLR.

\bibitem[{Jin et~al.(2023)Jin, Zhang, Zhang, Meng, Zhang, Zhu, and Han}]{jin2023patton}
Bowen Jin, Wentao Zhang, Yu~Zhang, Yu~Meng, Xinyang Zhang, Qi~Zhu, and Jiawei Han. 2023.
\newblock Patton: Language model pretraining on text-rich networks.
\newblock In \emph{Proceedings of the 61st Annual Meeting of the Association for Computational Linguistics (Volume 1: Long Papers)}, pages 7005--7020.

\bibitem[{Kipf and Welling(2016)}]{kipf2016semi}
Thomas~N Kipf and Max Welling. 2016.
\newblock Semi-supervised classification with graph convolutional networks.
\newblock \emph{arXiv preprint arXiv:1609.02907}.

\bibitem[{Lester et~al.(2021)Lester, Al-Rfou, and Constant}]{lester2021power}
Brian Lester, Rami Al-Rfou, and Noah Constant. 2021.
\newblock The power of scale for parameter-efficient prompt tuning.
\newblock \emph{arXiv preprint arXiv:2104.08691}.

\bibitem[{Li and Liang(2021)}]{li2021prefix}
Xiang~Lisa Li and Percy Liang. 2021.
\newblock Prefix-tuning: Optimizing continuous prompts for generation.
\newblock \emph{arXiv preprint arXiv:2101.00190}.

\bibitem[{Li et~al.(2023)Li, Li, Wang, Li, Sun, Cheng, and Yu}]{li2023survey}
Yuhan Li, Zhixun Li, Peisong Wang, Jia Li, Xiangguo Sun, Hong Cheng, and Jeffrey~Xu Yu. 2023.
\newblock A survey of graph meets large language model: Progress and future directions.
\newblock \emph{arXiv preprint arXiv:2311.12399}.

\bibitem[{Li et~al.(2024)Li, Li, Wang, Li, Sun, Cheng, and Yu}]{li_Survey_2024}
Yuhan Li, Zhixun Li, Peisong Wang, Jia Li, Xiangguo Sun, Hong Cheng, and Jeffrey~Xu Yu. 2024.
\newblock \href {https://arxiv.org/abs/2311.12399} {A {{Survey}} of {{Graph Meets Large Language Model}}: {{Progress}} and {{Future Directions}}}.
\newblock \emph{Preprint}, arXiv:2311.12399.

\bibitem[{Liu(2019)}]{liu2019roberta}
Yinhan Liu. 2019.
\newblock Roberta: A robustly optimized bert pretraining approach.
\newblock \emph{arXiv preprint arXiv:1907.11692}.

\bibitem[{Lu et~al.()Lu, Liu, Zhang, and Fu}]{luscale}
Jianglin Lu, Yixuan Liu, Yitian Zhang, and Yun Fu.
\newblock Scale-free graph-language models.
\newblock In \emph{The Thirteenth International Conference on Learning Representations}.

\bibitem[{Lu et~al.(2023)Lu, Xu, Wang, Bai, and Fu}]{lu2023latent}
Jianglin Lu, Yi~Xu, Huan Wang, Yue Bai, and Yun Fu. 2023.
\newblock Latent graph inference with limited supervision.
\newblock \emph{Advances in Neural Information Processing Systems}, 36:32521--32538.

\bibitem[{Ni et~al.(2019)Ni, Li, and McAuley}]{ni2019justifying}
Jianmo Ni, Jiacheng Li, and Julian McAuley. 2019.
\newblock Justifying recommendations using distantly-labeled reviews and fine-grained aspects.
\newblock In \emph{Proceedings of the 2019 conference on empirical methods in natural language processing and the 9th international joint conference on natural language processing (EMNLP-IJCNLP)}, pages 188--197.

\bibitem[{Nikolentzos et~al.(2020)Nikolentzos, Tixier, and Vazirgiannis}]{nikolentzos2020message}
Giannis Nikolentzos, Antoine Tixier, and Michalis Vazirgiannis. 2020.
\newblock Message passing attention networks for document understanding.
\newblock In \emph{Proceedings of the aaai conference on artificial intelligence}, volume~34, pages 8544--8551.

\bibitem[{Pennington et~al.(2014)Pennington, Socher, and Manning}]{pennington2014glove}
Jeffrey Pennington, Richard Socher, and Christopher~D Manning. 2014.
\newblock Glove: Global vectors for word representation.
\newblock In \emph{Proceedings of the 2014 conference on empirical methods in natural language processing (EMNLP)}, pages 1532--1543.

\bibitem[{Seo et~al.(2024)Seo, Kim, Yang, and Yang}]{seo_Unleashing_2024}
Hyunjin Seo, Taewon Kim, June~Yong Yang, and Eunho Yang. 2024.
\newblock \href {https://arxiv.org/abs/2405.18581} {Unleashing the {{Potential}} of {{Text-attributed Graphs}}: {{Automatic Relation Decomposition}} via {{Large Language Models}}}.
\newblock \emph{Preprint}, arXiv:2405.18581.

\bibitem[{Song et~al.(2024)Song, Mao, Xiao, Liu, Chen, Jin, Yang, Tang, and Liu}]{song2024pure}
Yu~Song, Haitao Mao, Jiachen Xiao, Jingzhe Liu, Zhikai Chen, Wei Jin, Carl Yang, Jiliang Tang, and Hui Liu. 2024.
\newblock A pure transformer pretraining framework on text-attributed graphs.
\newblock \emph{arXiv preprint arXiv:2406.13873}.

\bibitem[{Sun et~al.(2022)Sun, Dai, and Yu}]{sun2022does}
Ruoxi Sun, Hanjun Dai, and Adams~Wei Yu. 2022.
\newblock Does gnn pretraining help molecular representation?
\newblock \emph{Advances in Neural Information Processing Systems}, 35:12096--12109.

\bibitem[{Tan et~al.(2024)Tan, Lv, Huang, Zhang, Wang, and Yang}]{tan2024musegraph}
Yanchao Tan, Hang Lv, Xinyi Huang, Jiawei Zhang, Shiping Wang, and Carl Yang. 2024.
\newblock Musegraph: Graph-oriented instruction tuning of large language models for generic graph mining.
\newblock \emph{arXiv preprint arXiv:2403.04780}.

\bibitem[{Tang et~al.(2008)Tang, Zhang, Yao, Li, Zhang, and Su}]{tang2008arnetminer}
Jie Tang, Jing Zhang, Limin Yao, Juanzi Li, Li~Zhang, and Zhong Su. 2008.
\newblock Arnetminer: extraction and mining of academic social networks.
\newblock In \emph{Proceedings of the 14th ACM SIGKDD international conference on Knowledge discovery and data mining}, pages 990--998.

\bibitem[{Touvron et~al.(2023)Touvron, Lavril, Izacard, Martinet, Lachaux, Lacroix, Rozi{\`e}re, Goyal, Hambro, Azhar et~al.}]{touvron2023llama}
Hugo Touvron, Thibaut Lavril, Gautier Izacard, Xavier Martinet, Marie-Anne Lachaux, Timoth{\'e}e Lacroix, Baptiste Rozi{\`e}re, Naman Goyal, Eric Hambro, Faisal Azhar, and 1 others. 2023.
\newblock Llama: Open and efficient foundation language models.
\newblock \emph{arXiv preprint arXiv:2302.13971}.

\bibitem[{Veli{\v{c}}kovi{\'c} et~al.(2017)Veli{\v{c}}kovi{\'c}, Cucurull, Casanova, Romero, Lio, and Bengio}]{velivckovic2017graph}
Petar Veli{\v{c}}kovi{\'c}, Guillem Cucurull, Arantxa Casanova, Adriana Romero, Pietro Lio, and Yoshua Bengio. 2017.
\newblock Graph attention networks.
\newblock \emph{arXiv preprint arXiv:1710.10903}.

\bibitem[{Yang et~al.(2021)Yang, Liu, Xiao, Li, Lian, Agrawal, Singh, Sun, and Xie}]{yang2021graphformers}
Junhan Yang, Zheng Liu, Shitao Xiao, Chaozhuo Li, Defu Lian, Sanjay Agrawal, Amit Singh, Guangzhong Sun, and Xing Xie. 2021.
\newblock Graphformers: Gnn-nested transformers for representation learning on textual graph.
\newblock \emph{Advances in Neural Information Processing Systems}, 34:28798--28810.

\bibitem[{Ye et~al.(2024)Ye, Zhang, Wang, Xu, and Zhang}]{ye2024language}
Ruosong Ye, Caiqi Zhang, Runhui Wang, Shuyuan Xu, and Yongfeng Zhang. 2024.
\newblock Language is all a graph needs.
\newblock In \emph{Findings of the Association for Computational Linguistics: EACL 2024}, pages 1955--1973.

\bibitem[{Zhang et~al.(2025)Zhang, Chen, Bei, Yuan, Zhou, Hong, Dong, Chen, Chang, and Huang}]{zhang2025survey}
Qinggang Zhang, Shengyuan Chen, Yuanchen Bei, Zheng Yuan, Huachi Zhou, Zijin Hong, Junnan Dong, Hao Chen, Yi~Chang, and Xiao Huang. 2025.
\newblock A survey of graph retrieval-augmented generation for customized large language models.
\newblock \emph{arXiv preprint arXiv:2501.13958}.

\bibitem[{Zhang et~al.(2010)Zhang, Jin, and Zhou}]{zhang2010understanding}
Yin Zhang, Rong Jin, and Zhi-Hua Zhou. 2010.
\newblock Understanding bag-of-words model: a statistical framework.
\newblock \emph{International journal of machine learning and cybernetics}, 1:43--52.

\bibitem[{Zhao et~al.(2022)Zhao, Qu, Li, Yan, Liu, Li, Xie, and Tang}]{zhao2022learning}
Jianan Zhao, Meng Qu, Chaozhuo Li, Hao Yan, Qian Liu, Rui Li, Xing Xie, and Jian Tang. 2022.
\newblock Learning on large-scale text-attributed graphs via variational inference.
\newblock \emph{arXiv preprint arXiv:2210.14709}.

\bibitem[{Zhao et~al.(2023)Zhao, Zhou, Li, Tang, Wang, Hou, Min, Zhang, Zhang, Dong et~al.}]{zhao2023survey}
Wayne~Xin Zhao, Kun Zhou, Junyi Li, Tianyi Tang, Xiaolei Wang, Yupeng Hou, Yingqian Min, Beichen Zhang, Junjie Zhang, Zican Dong, and 1 others. 2023.
\newblock A survey of large language models.
\newblock \emph{arXiv preprint arXiv:2303.18223}.

\bibitem[{Zhou et~al.(2024{\natexlab{a}})Zhou, Dong, Huang, Liu, Zhou, and Xu}]{zhou2024quest}
Chuang Zhou, Junnan Dong, Xiao Huang, Zirui Liu, Kaixiong Zhou, and Zhaozhuo Xu. 2024{\natexlab{a}}.
\newblock Quest: Efficient extreme multi-label text classification with large language models on commodity hardware.
\newblock In \emph{Findings of the Association for Computational Linguistics: EMNLP 2024}, pages 3929--3940.

\bibitem[{Zhou et~al.(2023)Zhou, Chen, Dong, Zha, Zhou, and Huang}]{zhou2023adaptive}
Huachi Zhou, Hao Chen, Junnan Dong, Daochen Zha, Chuang Zhou, and Xiao Huang. 2023.
\newblock Adaptive popularity debiasing aggregator for graph collaborative filtering.
\newblock In \emph{Proceedings of the 46th international ACM SIGIR conference on research and development in information retrieval}, pages 7--17.

\bibitem[{Zhou et~al.(2024{\natexlab{b}})Zhou, Zhou, Chen, Liu, Yang, and Huang}]{zhou2024enhancing}
Huachi Zhou, Shuang Zhou, Hao Chen, Ninghao Liu, Fan Yang, and Xiao Huang. 2024{\natexlab{b}}.
\newblock Enhancing explainable rating prediction through annotated macro concepts.
\newblock In \emph{Proceedings of the 62nd Annual Meeting of the Association for Computational Linguistics (Volume 1: Long Papers)}, pages 11736--11748.

\bibitem[{Zhu et~al.(2024)Zhu, Wang, Shi, and Tang}]{zhu_Efficient_2024}
Yun Zhu, Yaoke Wang, Haizhou Shi, and Siliang Tang. 2024.
\newblock \href {https://arxiv.org/abs/2401.15569} {Efficient {{Tuning}} and {{Inference}} for {{Large Language Models}} on {{Textual Graphs}}}.
\newblock \emph{Preprint}, arXiv:2401.15569.

\end{thebibliography}

\end{document}